\documentclass{article}

 \usepackage[final,nonatbib]{neurips_2019}

\usepackage[utf8]{inputenc} 
\usepackage[T1]{fontenc}    
\usepackage{hyperref}       
\usepackage{url}            
\usepackage{booktabs}       
\usepackage{amsfonts}       
\usepackage{nicefrac}       
\usepackage{microtype}      
\usepackage{array}
\usepackage{tabulary}
\newcolumntype{K}[1]{>{\centering\arraybackslash}p{#1}}
\usepackage{siunitx}

\usepackage{xspace}
\usepackage{amsmath}
\usepackage{mathtools}
\usepackage{graphicx}
\graphicspath{ {./images/} }
\usepackage{lipsum}
\usepackage{graphicx}
\usepackage{comment}
\usepackage{amsfonts}
\usepackage{amssymb}
\usepackage{flushend} 
\usepackage{xcolor}
\usepackage{subcaption}
\usepackage{amssymb,amsmath,amsthm,enumitem}

\renewenvironment{thebibliography}[1]{
  \begin{oldthebibliography}{#1}
    \setlength{\itemsep}{0.8em}
    \setlength{\parskip}{0em}
}
{
  \end{oldthebibliography}
}

\title{Real-time Policy Distillation in\\ Deep Reinforcement Learning}

\author{
  Yuxiang Sun \\
  Computer Science and Engineering Department\\
  University of South Carolina\\
  \texttt{syuxiang@email.sc.edu} \\
  \And
  Pooyan Fazli\\
  Department of Computer Science\\
  San Francisco State University\\
  \texttt{pooyan@sfsu.edu} \\
}

\begin{document}

\maketitle

\begin{abstract}

Policy distillation in deep reinforcement learning provides an effective way to transfer control policies from a larger network to a smaller untrained network without a significant degradation in performance. However, policy distillation is underexplored in deep reinforcement learning, and existing approaches are computationally inefficient, resulting in a long distillation time. In addition, the effectiveness of the distillation process is still limited to the model capacity. We propose a new distillation mechanism, called \textit{real-time policy distillation}, in which training the \textit{teacher} model and distilling the policy to the \textit{student} model occur simultaneously. Accordingly, the teacher’s latest policy is transferred to the student model in real time. This reduces the distillation time to half the original time or even less and also makes it possible for extremely small student models to learn skills at the expert level. We evaluated the proposed algorithm in the Atari 2600 domain. The results show that our approach can achieve full distillation in most games, even with compression ratios up to $1.7\%$.

\end{abstract}

\section{Introduction}

Recent research has demonstrated the power of reinforcement learning to learn effective control policies in various domains. One breakthrough is the deep Q-network (DQN) \cite{mnih2015human}. In this approach, applied to the Atari domain, consecutive screenshots of games are fed as input to the network, and the network is expected to predict the best action that maximizes the expected future reward. Using this approach, a well-trained agent can outperform human-expert players in most Atari games. However, DQN models are usually cumbersome, which prevents their applications to compute- and power-constrained devices, such as drones and cell phones, that require efficient networks. To this end, policy distillation techniques have been widely investigated \cite{rusu2015policy,yin2017knowledge,parisotto2015actor} for their ability to transfer a policy to a compressed model with acceptable performance loss, primarily using distribution regression. However, most existing policy distillation methods require a long time to complete the distillation process, fail in some task domains or in providing full distillation, and have an approximation capacity that limits the model compression ratio.

In this work, we mainly focus on the quality of policy distillation and address the challenges of time and approximation capacity. First, we propose a new policy distillation method, in which training the teacher and distilling the policy to the student are carried out simultaneously in real time. This method is different from typical policy distillation approaches \cite{rusu2015policy}, in which the teacher model is trained, and then the student is trained via distillation under the supervision of the teacher model. Second, instead of optimizing the student’s policy only based on the output distribution regression between the teacher and student models \cite{rusu2015policy}, we import a \textit{self-learning} term in the loss function to help regulate the student’s behavior in an additional way. Third, beyond the supervision via distribution regression, we employ another channel of supervision by asking the student to imitate how the teacher sets reasonable targets. Finally, we investigate different variations of the loss function and their effects on the quality of the distillation process.

\section{Real-time Policy Distillation}

In policy distillation, one or more deep reinforcement learning models are trained as expert models that can provide supervision to student training. Distillation techniques such as prediction regression \cite{hinton2015distilling} and feature regression \cite{ahn2019variational,romero2014fitnets} have been comprehensively explored in different aspects of image classification. In deep reinforcement learning, KL divergence has been proved to be the most effective method to date \cite{rusu2015policy}. It is formulated as

\begin{equation} \label{eq:eq3}
{ \mathcal{L} }_{ KL }({ \theta  }^{ (S) })=\sum _{ i=1 }^{ N }{ { softmax(\frac { { q }_{ i }^{ (T) } }{ \tau  } ) }ln\left( \frac { { softmax(\frac { { q }_{ i }^{ (T) } }{ \tau  } ) } }{ { softmax(\frac { { q }_{ i }^{ (S) } }{ \tau  } ) } }  \right) , }   
\end{equation}
\vspace{-1ex}

where ${ { q }_{ i }^{ T } }$ and ${ { q }_{ i }^{ S } }$  are $q$-values approximated by the teacher model and the student model, respectively. ${ \theta  }^{ S }$ represents weights in the student model, and $N$ is the number of actions. The softmax function is used to normalize the $q$-values that can be taken as a probability distribution or confidence of the actions. The temperature $\tau$ is used to adjust the degree of similarity of the distributions that is expected to be achieved. By the definition of KL divergence, the teacher’s distribution is assumed to be the true distribution and used to supervise the student.

In the real-time distillation framework, both student and teacher models use the exact same architecture of the DQN. They both maintain two identical deep networks, a target network and a prediction network. Only the teacher interacts with the environment and determines the action to take, and the corresponding experiences are stored into a replay buffer. For each training iteration, the teacher and student models share the same data sampled from the replay buffer as a training set, and their weights are updated simultaneously and independently. The teacher’s policy is optimized by minimizing the loss of the Bellman error \cite{mnih2015human}, which we call the DQN loss:

\vspace{-1ex}
\begin{equation}\label{eq:eq2}
{ \mathcal{L} }_{ DQN }={ { { \mathbb{E} } } }_{ { s },a }{ [\underbrace { r_{ i+1 }+\gamma \times \underset { a }{ max } Q(s_{ i+1 },a,{ \theta  }_{ i }^{ - }) }_{ target } -\underbrace { Q(s_{ i },a;{ \theta  }_{ i }) }_{ prediction } ] }^{ 2 }.
\end{equation}
\vspace{-1ex}

In contrast, the student's loss is a combination of the DQN loss and the KL divergence:

\vspace{-1ex}
\begin{equation} \label{eq5}
{ \mathcal{L} }_{ student }={ \mathcal{L} }_{ KL }+{ \mathcal{L} }_{ DQN }.
\end{equation}
\vspace{-1ex}

We name the DQN loss in the student’s loss function as self-learning, as it encourages the student to find the most suitable way of shaping itself instead of strictly imitating the experts’ behavior, as even a very small model still has some approximating ability. When distillation is not informative enough or lacks high-quality supervision, applying knowledge only via the KL divergence channel would lead to a less effective distillation. In image classification \cite{jin2019knowledge,goldblum2019adversarially}, a combination of the student’s own classification and the distillation loss provides a reference from different aspects. In our case, according to our observation, the student’s weight distribution changes more drastically than that of the teacher during the training process. As the student adjusts itself to handle various states, self-learning is assumed to identify the right direction for updating the weights. Even when informative supervision is available, self-learning can still provide additional information, similar to self-adaptive adjustment or correctness.

\subsection{Imitation via Target-term}

The target term in the teacher’s DQN loss is decoupled as

\vspace{-1ex}
\begin{equation} \label{eq:eq4}
{ Y }_{ i }^{ (T) }=r_{ i+1 }+\gamma \times \underset { a }{ max } { Q }^{ (T) }(s_{ i+1 },a;{ \theta  }_{ i }^{ (T),- }).
\end{equation}
\vspace{-1ex}

Here, maximization is used to determine the action as well as estimate the $q$-value using the same approximator, that is, the target network. Inspired by the theoretical analysis in Double-DQN \cite{van2016deep,sutton2018reinforcement}, overestimation of the action value is more likely to occur in an insufficiently flexible approximation and may blow up via accumulation during training. In this prior work, double approximators were used to eliminate this penalty. Intuitively, the student’s self-learning accumulates more errors than a teacher model during training, so it is necessary to incorporate double estimators to reduce it. Therefore, inviting the teacher’s target network to be the second estimator in the student model that decides which action to take is a valid option. With this, the student’s target term in its DQN loss is changed to be

\vspace{-1ex}
\begin{equation} \label{eq:eq6}
{ Y }_{ i }^{ (S) }\equiv r_{ i+1 }+\gamma \times { Q }^{ (S) }(s_{ i+1 },\underset { a }{ argmax } { Q }^{ (T) }(s_{ i+1 },a;{ \theta  }_{ i }^{ (T),- });{ \theta  }_{ i }^{ (S),- }).
\end{equation}
\vspace{-1ex}

Note that the \textit{argmax} function extracts the action chosen by the teacher. In other words, the student uses weights ${ \theta  }_{ i }^{ (T),- }$ in the teacher's target network to determine the action and uses its own weights ${ \theta  }_{ i }^{ (S),- }$ to estimate the $q$-values. The student is required to imitate the teacher’s choices in the target estimation, a method that we call \textit{imitation}. Conversely, if the student takes the form of Equation \ref{eq:eq4}, it is called \textit{no-imitation}. Imitation via the target is a second channel of knowledge transference, beyond KL divergence.

\subsection{Forward KL vs. Reverse KL}

The uncertainty in action prediction is significant in some tasks, even when the agent has mastered some real skills. Consider the following extreme case, in which all actions are selected with equal probability both in the teacher and student models. The KL divergence tries to match the two random distributions. To deal with the uncertainty problem in image classification, Malinin and Gales \cite{malinin2019reverse} adopted reverse KL divergence (RKL) instead of forward KL divergence (FKL) when given highly uncertain data in classification. Back to reinforcement learning, the prediction uncertainty is generally high and causes more random action choices, which increases data uncertainty and confuses the student further. The reason is that the student learns from the teacher’s demonstration too. RKL originates from the posterior belief \cite{kingma2013auto} and can be decomposed to reverse cross-entropy (RCE) and self-entropy:

\vspace{-3ex}
\begin{equation}\label{eq:eq8}
\begin{split}
{ \mathcal{L} }_{ RKL }({ \theta  }^{ (S) })=\underbrace { \sum _{ i=1 }^{ N }{ { { -p } }_{ i }^{ (S) } } ln({ { p } }_{ i }^{ (T) }) }_{ RCE } -\underbrace { \sum _{ i=1 }^{ N }{ { -p }_{ i }^{ (S) }ln({ { { p } } }_{ i }^{ (S) }) }  }_{ Entropy } , 
\end{split}
\end{equation}
\vspace{-1ex}

where $p$ is the probability over actions, and $N$ is the number of actions. If there is no distinction between $q$-values, all actions are approximately equiprobable and selected with a uniform probability of $1/N$. The gradient of RKL for a single logit is

\vspace{-1ex}
\begin{equation}\label{eq:eq11}
\frac { \partial RKL }{ \partial { q }_{ i }^{ (S) } } =\frac { 1 }{ \tau  } (\frac { 1 }{ N } -\underbrace { ({ -p }_{ i }^{ (S) }ln{ (p }_{ i }^{ (S) }) }_{ Entropy } )).
\end{equation}
\vspace{-1ex}

According to Equation \ref{eq:eq11}, the gradient update only depends on the changes in the student’s entropy,
ignoring the teacher’s supervision. This mechanism can effectively prevent \textit{harmful} knowledge transfer via distillation, because if the teacher’s prediction ability degrades catastrophically and no constructive supervision is available, then the student is less affected by the teacher’s predictions, thus stabilizing the student’s learning. Theoretically \cite{malinin2019reverse}, in contrast to forward KL divergence, which is \textit{zero-avoiding} and drives the approximated distribution to be spread over each action, RKL is \textit{zero forcing},  which allows students to allocate zero probability to other actions as long as the student is confident with the predicted best action, consequently reducing the effect of the teacher.

However, uncertainty does not always damage learning effectiveness. According to our empirical study, the games that are more aware of state values, such as Space Invaders, are less affected by the application of FKL or RKL. This means that any action may yield a similar reward, so there is little impact due to prediction uncertainty. For games like Ms. Pacman, where each action choice determines the outcome, policies are more effectively distilled via RKL if high uncertainty exists.

\section{Experiments and Results}

To assess the performance of the proposed distillation approach, we selected seven games, one teacher model, Teacher=$\{32,64,64,512\}$, and three student models  \cite{rusu2015policy,yin2017knowledge} with different capacities, Net1=$\{ 16,32,32,256\}$, Net2=$\{16,16,16,128\}$, and Net3=$\{16,16,16,64\}$. The first three numbers represent the number of filters in the convolutional layers, and the last is the number of neurons in a fully connected layer. Students’ corresponding compression ratios, in terms of the total number of parameters, are $25.2\%$, $6.7\%$, and $3.7\%$. To save training time, in each experiment, one teacher model and multiple student models with different sizes are constructed in the same graph, and all students share the same teacher. All training and evaluation settings are exactly the same as in DQN \cite{mnih2015human} except for the option of \textit{null-op}. Ideally, after the teacher’s convergence, additional iterations may help the student to consolidate accomplishments, but when training time is limited, we force all models to stop at the same epoch. Note that we have much fewer training epochs than does Policy Distillation (PD) \cite{rusu2015policy}, only as a rough reference, and the average training time is $200$ hours for each experiment.

\newcommand{\ra}[1]{\renewcommand{\arraystretch}{#1}}

\begin{table*}
\centering
\caption{Training results by applying the \textit{imitation} method for different model sizes. Scores calculated by the percentage of teacher's scores (\%). RTPD: Real-time Policy Distillation, PD: Policy Distillation \cite{rusu2015policy}. Net1, Net2, and Net3 are the student models.}
\ra{1.25}
\begin{tabular}{@{}lrrrcrrrcrrr@{}} 
\cmidrule[\heavyrulewidth](lr){2-10}

& \multicolumn{3}{c}{$Net1$} & \multicolumn{3}{c}{$Net2$} & \multicolumn{3}{c}{$Net3$}\\

& \multicolumn{1}{c}{$PD$} & \multicolumn{2}{c}{$RTPD$} &
\multicolumn{1}{c}{$PD$}& \multicolumn{2}{c}{$RTPD$} & \multicolumn{1}{c}{$PD$} & \multicolumn{2}{c}{$RTPD$}\\

\cmidrule(lr){2-2} \cmidrule{3-4} \cmidrule(lr){5-5} \cmidrule{6-7} \cmidrule(lr){8-8} \cmidrule{9-10}
& $Max$ & $Max$ & $Mean$ & $Max$ & $Max$ & $Mean$ & $Max$ & $Max$ & $Mean$\\ \midrule
$FKL$\\
\textbf{Ms.Pacman} &102.5&95.2&\textbf{112.0}&86.4&96.9&\textbf{92.3}&\textbf{96.2}&97.7&58.2\\
\textbf{Qbert} &129.6&80.8&\textbf{234.0}&130.6&76.5&\textbf{320.7}&\textbf{107.9}&55.5&101.9\\
\textbf{Beamrider} &\textbf{87.1}&47.4&35.8&\textbf{85.3}&62.6&37.0&\textbf{75.2}&47.6&15.91\\
\textbf{Pong} &\textbf{100.6}&104.8&97.5&\textbf{103.7}&98.0&100.7&96.9&100.8&\textbf{99.1}\\
\textbf{SpaceInvaders} &78.6&101.9&\textbf{95.1}&49.4&100.6&\textbf{100.5}&20.9&94.2&\textbf{95.4}\\
\textbf{Breakout} &\textbf{105.6}&73.8&100.1&\textbf{98.1}&91.5&97.3&78.6&69.5&\textbf{91.1}\\
\textbf{Enduro} &\textbf{142.4}&103.9&116.4&\textbf{141.3}&106.5&105.1&117.1&109.2&\textbf{120.3}\\
\midrule
$RKL$\\
\textbf{Ms.Pacman} &102.5&130.5&\textbf{112.9}&86.4&120.3&\textbf{111.5}&96.2&124.0&\textbf{99.2} \\
\textbf{Qbert} &129.6&91.8&\textbf{306.7}&130.6&86.6&\textbf{229.3}&107.9&75.6&\textbf{233.7}\\
\textbf{Beamrider} &87.1&93.0&\textbf{108.6}& \textbf{85.3}&79.0&83.6&\textbf{75.2}&60.1&59.1\\
\textbf{Pong} &100.6&105.9&\textbf{113.3}&103.7&104.6& \textbf{114.4}&96.9&101.8&\textbf{100.5}\\
\textbf{SpaceInvaders} &78.6&108.9&\textbf{93.7}&49.4&105.3&\textbf{97.4}&20.9&102.6&\textbf{98.4}\\
\textbf{Breakout} &105.6&102.9&\textbf{106.9}&98.1&99.5&\textbf{101.3}&78.6&101.2&\textbf{107.7}\\
\textbf{Enduro} &\textbf{142.4}&99.9&104.0&\textbf{141.3}&101.4&104.3&\textbf{117.1}&100.6&103.6\\

\bottomrule
\end{tabular}
\label{tab:all}
\end{table*}

If perfect distillation performance of our algorithm is achieved, we expect the student’s learning pace to keep up with the teacher’s in real time and the student to outperform the teacher with respect to maximum and mean scores. All models are evaluated sequentially every \num[group-separator={,}]{25000} updates (called epochs) by playing up to $30$ episodes. Limited by computation resources, we only consider $100$ epochs. Besides comparing the student’s maximum score over the whole training as a percentage of the teacher’s maximum score, we also calculate the mean of all score percentages in the last $10$ epochs.

In Table \ref{tab:all}, we compare the mean percentages of our proposed method with \textit{Policy Distillation} (PD) \cite{rusu2015policy}, as the mean value is more eligible to reflect the distillation performance. It is observed that the real-time architecture, combined with the imitation method, guarantees almost $100\%$ distillation for most games, especially in terms of the mean value. Also, the distillation is less affected by the model capacity, and even the smallest model, Net3, is still able to achieve a performance comparable to Net1. We also conducted experiments applying RKL in contrast to FKL, which verifies the benefit of RKL.

\mathchardef\mhyphen="2D
\begin{table*}[t!]
\centering
\caption{Comparison of \textit{imitation} and \textit{no-imitation} with FKL (\% of teacher).}
\ra{1.25}
\begin{tabular}{@{}lrrrcrrrcrrr@{}}
\cmidrule[\heavyrulewidth](lr){2-9}

& \multicolumn{4}{c}{$Net4$} & \multicolumn{4}{c}{$Net5$} \\
& \multicolumn{2}{c}{$Imitation$} & \multicolumn{2}{c}{$No \mhyphen imitation$} &
\multicolumn{2}{c}{$Imitation$}& \multicolumn{2}{c}{$No \mhyphen imitation$} \\

\cmidrule(lr){2-3} \cmidrule{4-5} \cmidrule(lr){6-7} \cmidrule{8-9} 
& $Max$ & $Mean$ & $Max$ & $Mean$ & $Max$ & $Mean$ & $Max$ & $Mean$\\ \midrule
\textbf{Ms.Pacman}&\textbf{91.7}&\textbf{74.8}&81.0&63.9&\textbf{96.0}&\textbf{79.1}&84.4&73.7\\
\textbf{SpaceInvaders}&\textbf{99.8}&\textbf{116.4}&94.9&106.6&\textbf{94.2}&86.0&87.9&\textbf{89.3}\\
\textbf{Breakout}&\textbf{96.3}&102.2&81.0&\textbf{103.0}&\textbf{92.1}&\textbf{100.4}&87.5&100.0\\

\bottomrule
\end{tabular}
\label{tab:imitation}
\end{table*}

In Table \ref{tab:imitation}, two extra even smaller models, Net4=$\{8,8,16,64\}$ and Net5=$\{8,16,8,64\}$ were added to assess the advantage of imitation in very small models. The compression ratios are $3.2\%$ and $1.7\%$, respectively. The result shows the imitation method truly helps improve the overall distillation quality for the three games.

\section{Conclusion}

Our real-time distillation architecture reduces the distillation time significantly and transfers knowledge to extremely small models without much degradation in performance. The investigation of the imitation method and two forms of KL divergence may contribute some insights into the improvement of distillation quality and other reinforcement learning tasks, such as robot control and planning.

\bibliographystyle{plain}
\bibliography{main}

\begin{thebibliography}{10}

\bibitem{ahn2019variational}
Sungsoo Ahn, Shell~Xu Hu, Andreas Damianou, Neil~D Lawrence, and Zhenwen Dai.
\newblock Variational information distillation for knowledge transfer.
\newblock In {\em Proceedings of the IEEE Conference on Computer Vision and
  Pattern Recognition (CVPR)}, pages 9163--9171, 2019.

\bibitem{goldblum2019adversarially}
Micah Goldblum, Liam Fowl, Soheil Feizi, and Tom Goldstein.
\newblock Adversarially robust distillation.
\newblock {\em arXiv:1905.09747}, 2019.

\bibitem{hinton2015distilling}
Geoffrey Hinton, Oriol Vinyals, and Jeff Dean.
\newblock Distilling the knowledge in a neural network.
\newblock {\em arXiv:1503.02531}, 2015.

\bibitem{jin2019knowledge}
Xiao Jin, Baoyun Peng, Yichao Wu, Yu~Liu, Jiaheng Liu, Ding Liang, Junjie Yan,
  and Xiaolin Hu.
\newblock Knowledge distillation via route constrained optimization.
\newblock {\em arXiv:1904.09149}, 2019.

\bibitem{kingma2013auto}
Diederik~P Kingma and Max Welling.
\newblock Auto-encoding variational bayes.
\newblock {\em arXiv:1312.6114}, 2013.

\bibitem{malinin2019reverse}
Andrey Malinin and Mark Gales.
\newblock Reverse kl-divergence training of prior networks: Improved
  uncertainty and adversarial robustness.
\newblock {\em arXiv:1905.13472}, 2019.

\bibitem{mnih2015human}
Volodymyr Mnih, Koray Kavukcuoglu, David Silver, Andrei~A Rusu, Joel Veness,
  Marc~G Bellemare, Alex Graves, Martin Riedmiller, Andreas~K Fidjeland, Georg
  Ostrovski, et~al.
\newblock Human-level control through deep reinforcement learning.
\newblock {\em Nature}, 518(7540):529, 2015.

\bibitem{parisotto2015actor}
Emilio Parisotto, Jimmy~Lei Ba, and Ruslan Salakhutdinov.
\newblock Actor-mimic: Deep multitask and transfer reinforcement learning.
\newblock {\em arXiv:1511.06342}, 2015.

\bibitem{romero2014fitnets}
Adriana Romero, Nicolas Ballas, Samira~Ebrahimi Kahou, Antoine Chassang, Carlo
  Gatta, and Yoshua Bengio.
\newblock Fitnets: Hints for thin deep nets.
\newblock {\em arXiv:1412.6550}, 2014.

\bibitem{rusu2015policy}
Andrei~A Rusu, Sergio~Gomez Colmenarejo, Caglar Gulcehre, Guillaume Desjardins,
  James Kirkpatrick, Razvan Pascanu, Volodymyr Mnih, Koray Kavukcuoglu, and
  Raia Hadsell.
\newblock Policy distillation.
\newblock {\em arXiv:1511.06295}, 2015.

\bibitem{sutton2018reinforcement}
Richard~S Sutton and Andrew~G Barto.
\newblock {\em Reinforcement learning: An introduction}.
\newblock MIT press, 2018.

\bibitem{van2016deep}
Hado Van~Hasselt, Arthur Guez, and David Silver.
\newblock Deep reinforcement learning with double q-learning.
\newblock In {\em Proceedings of the Thirtieth AAAI Conference on Artificial
  Intelligence (AAAI)}, pages 2094--2100, 2016.

\bibitem{yin2017knowledge}
Haiyan Yin and Sinno~Jialin Pan.
\newblock Knowledge transfer for deep reinforcement learning with hierarchical
  experience replay.
\newblock In {\em Proceedings of the Thirty-First AAAI Conference on Artificial
  Intelligence (AAAI)}, pages 1640--1646, 2017.

\end{thebibliography}

\end{document}